\documentclass[a4paper]{article}
\usepackage{graphics} 
\usepackage{mathptmx} 
 \usepackage{graphicx}
\usepackage{color,colortbl}
\usepackage{xcolor}
\usepackage{graphicx}
\usepackage{longtable}
\usepackage{verbatim}
\usepackage{tikz}
\usepackage{tikz-qtree}
\usetikzlibrary{shapes,shapes.geometric,arrows,fit,calc,positioning}
\usepackage{amsmath}
\usepackage{listings}
\definecolor{lightGray}{gray}{0.9}

\newtheorem{example}{Example}
\newtheorem{pattern}{Antipattern}
\newcommand{\co}{}
\newcommand{\cov}{\text{Covid-19} }

\newcommand{\quant}{q: }
\newcommand{\fred}{FRED }

\begin{document}


\title{Detecting fake news for the new coronavirus \\
by reasoning on the \cov ontology}


\author{Adrian Groza\\
Computer Science Department, Technical University of Cluj-Napoca,\\
Memorandumului 14, Cluj-Napoca, Romania\\
\texttt{adrian.groza@cs.utcluj.ro}
}


\maketitle

\begin{abstract}
In the context of the \cov pandemic, many were quick to spread deceptive information~\cite{cinelli2020covid}. 
I investigate here how reasoning in Description Logics (DLs) can detect inconsistencies between trusted medical sources and not trusted ones.
The not-trusted information comes in natural language (e.g. "\cov affects only the elderly"). 
To automatically convert into DLs, I used the FRED converter~\cite{gangemi2017semantic}. 
Reasoning in Description Logics is then performed with the Racer tool~\cite{haarslev2012racerpro}.
\end{abstract}



\section{Introduction}
In the context of \cov pandemic, many were quick to spread deceptive information~\cite{cinelli2020covid}. 
Fighting against misinformation requires tools from various domains like law, education, and also from information technology~\cite{vosoughi2018spread, lazer2018science}.

Since there is a lot of trusted medical knowledge already formalised, I investigate here how an ontology on \cov could be used to signal fake news.

I investigate here how reasoning in description logic can detect inconsistencies between a trusted medical source and not trusted ones.
The not-trusted information comes in natural language (e.g. "\cov affects only elderly"). 
To automatically convert into description logic (DL), I used the FRED converter~\cite{gangemi2017semantic}. 
Reasoning in Description Logics is then performed with the 
Racer reasoner~\cite{haarslev2012racerpro}\footnote{The Python sources and the formalisation in Description Logics (KRSS syntax) are available at https://github.com/APGroza/ontologies}

The rest of the paper is organised as follows: 
Section~\ref{sec:dl} succinctly introduces the syntax of description logic and shows how inconsistency can be detected by reasoning. 
Section~\ref{sec:fred} analyses \fred translations for the \cov myths.
Section~\ref{sec:patterns} illustrates how to formalise knowledge patterns for automatic conflict detection.
Section~\ref{sec:related} browses related work, while section~\ref{sec:conclusion} concludes the paper.

\section{Finding inconsistencies using Description Logics}
\label{sec:dl}

\subsection{Description Logics}
\label{subsec:descriptionLogic}
In the Description Logics, concepts are built using the set of constructors formed by negation, conjunction, disjunction, value restriction, and existential restriction~\cite{baader2003description} (Table~\ref{tab:dl}). 
Here, $C$ and $D$ represent concept descriptions, while $r$ is a role name.
The semantics is defined based on an interpretation $I = (\Delta^{I}, \cdotp^{I})$, where the domain $\Delta^{I}$ of $I$ contains a non-empty set of individuals, and the interpretation function $\mathit{\cdotp^I}$ maps each concept name $C$ to a set of individuals $C^I\in \Delta^I$ and each role $r$ to a binary relation $r^I\in \Delta^I \times \Delta^I$. 
The last column of Table~\ref{tab:dl} shows the extension of $\cdotp^{I}$ for non-atomic concepts.

\begin{table}
\centering
\begin{footnotesize}
\caption{Syntax and semantics of DL}
\label{tab:dl}
\begin{tabular}{lll}
\hline
{\it Constructor} & {\it Syntax} & {\it Semantics} \\
\hline
conjunction & $C \sqcap D$ & $C^I \cap D^I$\\ 
disjunction & $C \sqcup D$ & $C^I \cup D^I$\\ 
existential restriction & $\exists r.C$ & $\{x\in \Delta^I | \exists y: (x,y)\in r^I \wedge y\in C^I\}$ \\ 
value restriction & $\forall r.C$ & $\{x\in \Delta^I | \forall y: (x,y)\in r^I \rightarrow y\in C^I \}$ \\ 
individual assertion & $a:C$ & $\{a\} \in C^I$\\ 
role assertion & $r(a,b)$ & $(a,b)\in r^I$\\ \hline
\end{tabular}
\end{footnotesize}
\end{table}

A terminology $\mathit{TBox}$ is a finite set of terminological axioms of the forms C $\equiv$ D or C $\sqsubseteq$ D.

\begin{example}[Terminological box]
"Coronavirus disease (\cov) is an infectious disease caused by a newly discovered coronavirus" can be formalised as:
\begin{eqnarray}
\cov \equiv CoronavirusDisease\\
InfectiousDisease \sqsubseteq Disease \\ 
CoronavirusDisease \sqsubseteq InfectiosDisease\ \sqcap\ \forall causedBy.NewCoronavirus \label{eq:c3}
\end{eqnarray}
Here the concept $\cov$ is the same as the concept $CoronavirusDisease$. 
We know that an infectious disease is a disease (i.e. the concept $InfectiousDisease$ is included in the more general concept $Disease$). 
We also learn from~\eqref{eq:c3} that the coronovirus disease in included the intersection of two sets: the set $InfectionDisease$ and the set of individuals for which all the roles $causedBy$ points towards instances from the concept $NewCoronavirus$. 
\end{example}

An assertional box $\mathit{ABox}$ is a finite set of concept assertions $i:C$ or role assertions {\it r(i,j)}, where $C$ designates a concept, $r$ a role, and $i$ and $j$ are two individuals.
\begin{example}[Assertional Box]
$\text{SARS-CoV-2}:Virus$ says that the individual $\text{SARS-CoV-2}$ is an instance of the concept $Virus$. 
$hasSource(\text{SARS-CoV-2},bat)$ formalises the information that SARS-Cov-2 comes from the bats. Here the role 
$hasSource$ relates two individuals $\text{SARS-CoV-2}$  and $bat$ that is an instance of mammals (i.e. $bat:Mammal$). 
\end{example}

A concept $C$ is satisfied if there exists an interpretation $I$ such that $C^I \neq \emptyset$.  
The concept $D$ subsumes the concept $C$ ($\mathit{C\sqsubseteq D}$) if $\mathit{C^I \subseteq D^I}$ for all interpretations $I$.
Constraints on concepts (i.e. {\it disjoint}) or on roles 
({\it domain}, {\it range}, {\it inverse} role, or {\it transitive} properties) can be specified in more expressive description logics\footnote{I provide only some basic terminologies of description logics in this paper to make it self-contained. For a detailed explanation about families of Description Logics, the reader is referred to~\cite{baader2003description}.}.
By reasoning on this mathematical constraints, one can detect inconsistencies among different pieces of knowledge, as illustrated in the following inconsistency patterns. 

\subsection{Inconsistency patterns} 
An ontology $\mathcal{O}$ is incoherent iff there exists an unsatisfiable concept in $\mathcal{O}$.
\begin{example}[Incoherent ontology]
\begin{align}
\cov \sqsubseteq InfectionDisease\\
\cov \sqsubseteq \neg InfectionDisease
\end{align}
is incoherent because $COVID19$ is unsatisfiable in $O$ since it included to two disjoint sets. 
\end{example}

In most of the cases, reasoning is required to signal that a concept is includes in two disjoint concepts.

\begin{example}[Reasoning to detect incoherence]
\begin{eqnarray}
\cov \sqsubseteq InfectionDisease \label{eq:c4}\\
InfectiousDisease \sqsubseteq Disease\ \sqcap\ \exists causedBy. (Bacteria \sqcup Virus \sqcup Fungi \sqcup Parasites)\label{eq:c5}\\
\cov \sqsubseteq \neg Disease \label{eq:c6}
\end{eqnarray}
From axioms~\ref{eq:c4} and~\ref{eq:c5}, one can deduce that \cov is included in the concept $Disease$. 
From axiom~\ref{eq:c6}, one learns the opposite: \cov is outside the same set $Disease$. 
A reasoner on Description Logics will signal an incoherence. 
\end{example}

An ontology is inconsistent when an unsatisfiable concept is instantiated.
For instance, inconsistency occurs when the same individual is an instance of two disjoint concepts 
\begin{example}[Inconsistent ontology]
\begin{eqnarray}
\text{SARS-CoV-2}:Virus \label{eq:c7}\\
\text{SARS-CoV-2}:Bacteria \label{eq:c8}\\
Virus \sqsubseteq Bacteria \label{eq:c9}
\end{eqnarray}
We learn that SARS-CoV-2 is an instance of both $Virus$ and $Bacteria$ concepts. 
Axiom~\eqref{eq:c6} states the viruses are disjoint of bacteria. 
A reasoner on Description Logics will signal an inconsistency.
\end{example}

Two more examples of such antipatterns\footnote{There are more such antipatterns~\cite{roussey2013antipattern} that trigger both incoherence and inconsistency.} are: 

\begin{pattern}[Onlyness Is Loneliness - OIL]
\begin{eqnarray}
OIL_1: A\sqsubseteq \forall r. B\\
OIL_2: A\sqsubseteq \forall r. C\\
OIL_3: B \sqsubseteq \neg C
\end{eqnarray}
Here, concept $A$ can only be linked with role $r$ to $B$.
Next, $A$ can only be linked with role $r$ to $C$, disjoint with $B$. 
\end{pattern}

\begin{example}[OIL antipattern]
\begin{align}
OIL_1: Antibiotics\sqsubseteq \forall kills. Virus\\
OIL_2: Antibiotics\sqsubseteq \forall kills. Bacteria\\
OIL_3: Virus \sqsubseteq \neg Bacteria
\end{align}
\end{example}
   
\begin{pattern}[Universal Existence - UE]
\begin{align}
UE_1: A\sqsubseteq \forall r. C\\
UE_2: A\sqsubseteq \exists r. B\\
UE_3: B \sqsubseteq \neg C
\end{align}
Axiom $UE_2$ adds an existential restriction for the concept $A$ conflicting with the existence of an universal restriction for the same concept $A$ in $UE_1$. 
\end{pattern}

\begin{example}[UE antipattern]
\begin{align}
UE_1: Antibiotics\sqsubseteq \forall kills. Virus\\
UE_2: Antibiotics\sqsubseteq \exists kills. Bacteria\\
UE_3: Virus \sqsubseteq \neg Bacteria
\end{align}
\end{example}

Assume that axioms $UE_2$ and $UE_3$ comes from a trusted source, while axiom $UE_1$ from the social web. 
By combining all three axioms, a reasoner will signal the inconsistency or incoherence. 
The technical difficulty is that information from social web comes in natural language.

\section{Analysing misconceptions with \cov ontology}

\begin{table}
\begin{footnotesize}
\caption{Sample of myths versus facts on \cov\label{tab:myths}}
\begin{tabular}{lp{6.5cm}lp{6.5cm}} \hline
 & Myth & & Fact \\ \hline 
$m_1$ & 5G mobile networks spread \cov  & $f_1$& Viruses can not travel on radio waves/mobile networks\\
$m_2$ & Exposing yourself to the sun or to temperatures higher than 25C degrees prevents the coronavirus disease & $f_2$ & You can catch \cov, no matter how sunny or hot the weather is\\
$m_3$ & You can not recover from the coronavirus infection & $f_3$ & Most of the people who catch \cov can recover and eliminate the virus from their bodies.\\
$m_4$ & \cov can not be transmitted in areas with hot and humid climates & $f_4$ & \cov can be transmitted in all areas\\
$m_5$ & Drinking excessive amounts of water can flush out the virus & $f_5$ &  Drinking excessive amounts of water can not flush out the virus\\
$m_6$ & Regularly rinsing your nose with saline help prevent infection with \cov & $f_6$ & There is no evidence that regularly rinsing the nose with saline has protected people from infection with the new coronavirus \\ 
$m_7$ & Eating raw ginger counters the coronavirus & $f_7$ & There is no evidence that eating garlic has protected people from the new coronavirus \\
$m_9$ & The new coronavirus can be spread by Chinese food & $f_9$ & The new coronavirus can not be transmitted through food\\
$m_{10}$ & Hand dryers are effective in killing the new coronavirus & $f_{10}$ &  Hand dryers are not effective in killing the 2019-nCoV\\
$m_{11}$ & Cold weather and snow can kill the new coronavirus & $f_{11}$ & Cold weather and snow can not kill the new coronavirus\\
$m_{12}$ & Taking a hot bath prevents the new coronavirus disease & $f_{12}$ & Taking a hot bath will not prevent from catching \cov\\
$m_{13}$ & Ultraviolet disinfection lamp kills the new coronavirus & $f_{13}$ & Ultraviolet lamps should not be used to sterilize hands or other areas of skin as UV radiation can cause skin irritation\\
$m_{14}$ & Spraying alcohol or chlorine all over your body kills the new coronavirus & $f_{14}$ & Spraying alcohol or chlorine all over your body will not kill viruses that have already entered your body\\
$m_{15}$ & Vaccines against pneumonia protect against the new coronavirus & $f_{15}$& Vaccines against pneumonia, such as pneumococcal vaccine and Haemophilus influenza type B (Hib) vaccine, do not provide protection against the new coronavirus\\
$m_{16}$ & Antibiotics are effective in preventing and treating the new coronavirus & $f_{16}$ & Antibiotics do not work against viruses, only bacteria.\\
$m_{17}$ & High dose of Vitamin C heals \cov & $f_{17}$ & No supplement cures or prevents disease\\
$m_{19}$ & The pets transmit the Coronavirus to humans & $f_{19}$ & There are currently no reported cases of people catching the coronavirus from animals\\
$m_{22}$ & If you can't hold your breath for 10 seconds, you have a coronavirus disease & $f_{22}$ & You can not confirm coranovirus disease with breathing exercise\\
$m_{24}$ & Drinking alcohol prevents \cov & $f_{24}$ & Drinking alcohol does not protect against \cov and can be dangerous\\
$m_{27}$ & Eating raw lemon counters coronavirus & $f_{27}$ &  No food cures or prevents disease \\
$m_{29}$ & Zinc supplements can lower the risk of contracting Covid-19 & $f_{29}$ & No supplement cures or prevents disease \\
$m_{31}$ & Vaccines against flu protect against the new coronavirus & $f_{31}$ & Vaccines against flu do not protect against the new coronavirus\\
$m_{32}$ & The new coronavirus can be transmitted through mosquito& $f_{32}$& The new coronavirus can not be transmitted through mosquito\\
$m_{33}$ & \cov can affect elderly only & $f_{33}$ & \cov can affect anyone\\ \hline
\end{tabular}
\end{footnotesize}
\end{table}

Sample medical misconceptions on \cov are collected in Table~\ref{tab:myths}). 
Organisations such as WHO provides facts for some myths (denoted $f_i$ in the table).

Let for instance myth $m_1$ with the formalisation:
\begin{align}
5G: MobileNetwork \label{eq:m1_1}\\ 
covid19: Virus \label{eq:m1_2}\\
spread(5G,covid19)\label{eq:m1_3} 
\end{align}

Assume the following formalisation for the corresponding fact $f_1$:
\begin{align}
Virus  \sqsubseteq \neg (\exists travel.(RadioWaves\ \sqcup\ MobileNetworks)) 
\end{align}

The following line of reasoning signals that the ontology is inconsistent:
\begin{align}
    Virus  \sqsubseteq \neg (\exists travel.MobileNetworks)\\
    Virus  \sqsubseteq \forall travel.\neg MobileNetworks\\
    Virus  \sqsubseteq \forall spread. \neg MobileNetworks \label{eq:f1_3}
\end{align}
Here we need the subsumption relation between roles ($travel\sqsubseteq spread$). The reasoner finds that the individual $5G$ (which is a mobile network by axiom~\eqref{eq:m1_1})
that spreads $COVID19$ (which is a virus by axiom~\eqref{eq:m1_2}) is in conflict with the axiom~\eqref{eq:f1_3}. 

As a second example, let the myth $m_{33}$ in Table~\ref{tab:myths}:  
\begin{equation}
  \cov  \sqsubseteq \forall affects.Elderly 
\end{equation}
The corresponding fact $f_{33}$ states: 
\begin{equation}
  \cov  \sqsubseteq \forall affects.Person 
\end{equation}
The inconsistency will be detected on the Abox contains and individual affected by \cov and who is not elderly:
\begin{align}
affectedBy(jon,\cov)\\    
hasAge(jon,40)
\end{align}
We need also some background knowledge:
\begin{align}
Elderly  \sqsubseteq Person \sqcap (> hasAge\ 65)\\
affects^- \equiv affectedBy \\
\cov \equiv one-of(\cov)
\end{align}
Based on the definition of $Elderly$ and on jon's age, the reasoner learns that $jon$ does not belong to that concept (i.e $jon:\neg Elderly$).
From the inverse roles $affects^- \equiv affectedBy$, one learns that the virus \cov affects $jon$.
Since the concept \cov includes only the individual with the same name \cov (defined with the constructor $one-of$ for nominals), the reasoner will be able to detect inconsistency.

Note that we need some background knowledge (like definition of $Elderly$) to signal conflict. 
Note also the need of a trusted \cov ontology.

There is ongoing work on formalising knowledge about \cov. 
First,\textit{Coronavirus Infectious Disease Ontology} (CIDO) \footnote{http://bioportal.bioontology.org/ontologies/CIDO}.
Second, the Semantics for \cov Discovery\footnote{https://github.com/fhircat/CORD-19-on-FHIR} adds semantic annotations to the CORD-19 dataset. 
The CORD-19 dataset was obtained by automatically analysing publications on \cov. 

Note also that the above formalisation was manually obtained. 
Yet, in most of the cases we need automatic translation from natural language to description logic.

\section{Automatic conversion of the \cov myth into Description Logic with FRED}
\label{sec:fred}

Transforming unstructured text into a formal representation is an important task for the Semantic Web. 
Several tools are contributing towards this aim: FRED~\cite{gangemi2017semantic}, OpenEI~\cite{martinez2018openie}, controlled languages based approach (e.g. ACE), Framester~\cite{gangemi2016framester}, or KNEWS~\cite{alam2017event}. 
We here the FRED tool, that takes a text an natural language and outputs a formalisation in description logic.

\fred is a machine reader for the Semantic Web that relies on Discourse Representation Theory, 
Frame semantics and Ontology Design Patterns\footnote{http://ontologydesignpatterns.org/ont}~\cite{draicchio2013fred,gangemi2017semantic}.
FRED leverages multiple natural language processing (NLP) components by integrating their outputs into a unified result, which is formalised as an RDF/OWL graph.
Fred relies on several NLP knowledge resources (see Table ~\ref{fig:fredlist}). 
VerbNet~\cite{schuler2005verbnet} contains semantic roles and patterns that are structure into a taxonomy. 
 FrameNet~\cite{baker1998berkeley} introduces frames to describe a situation, state or action. 
 The elements of a frame include: agent, patient, time, location.
 A frame is usually expressed by verbs or other linguistic constructions, hence all occurrences of frames are formalised as OWL n-ary relations, all being instances of some type of event or situation.


We exemplify next, how \fred handles linked data, compositional semantics, plurals, modality and negations with examples related to \cov:

\subsection{Linked Data and compositional semantics}
 
 \begin{table*}
 \begin{footnotesize}
 \caption{FRED's knowledge resources and their prefixes used for the \cov myts ontology\label{fig:fredlist}}
\centering
\begin{tabular}{lll}\hline
Ontology & Prefix &  Name Space \\ \hline
\cov myths & covid19.m: & http://www.ontologydesignpatterns.org/ont/\cov/covid-19-myths.owl\#\\
VerbNet roles & vn.role: & http://www.ontologydesignpatterns.org/ont/vn/abox/role/\\
VerbNet concepts & vn.data:& http://www.ontologydesignpatterns.org/ont/vn/data/\\
FrameNet frame & ff: &http://www.ontologydesignpatterns.org/ont/framenet/abox/frame/\\
FrameNet element  &fe:& http://www.ontologydesignpatterns.org/ont/framenet/abox/fe/\\
DOLCE+DnS Ultra Light & dul: & http://www.ontologydesignpatterns.org/ont/dul/DUL.owl\#\\
WordNet & wn30: & http://www.w3.org/2006/03/wn/wn30/instances/\\
Boxer & boxer: & http://ontologydesignpatterns.org/ont/boxer/boxer.owl\#\\
Boxing & boxing: & http://ontologydesignpatterns.org/ont/boxer/boxing.owl\#\\
DBpedia & dbpedia:& http://dbpedia.org/resource/\\
schema.org& schemaorg: &  http://schema.org/ \\ 
Quantity & \quant & \\ \hline
\end{tabular}
\end{footnotesize}
\end{table*}

\begin{figure}
\begin{center}
\includegraphics[width=\textwidth]{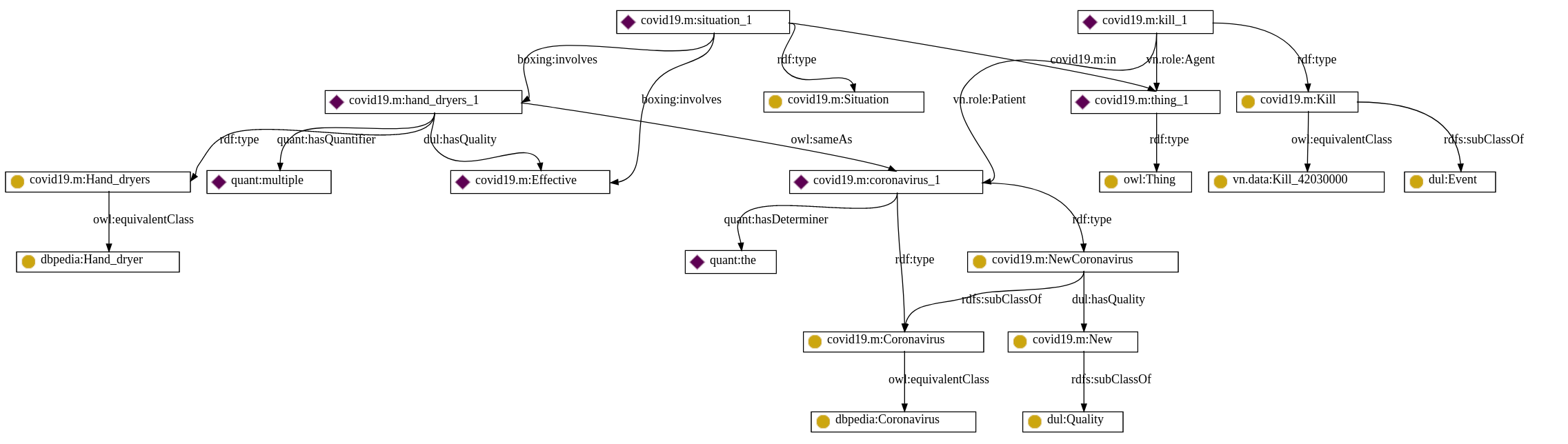}    
\end{center}
\caption{Translating the myth: "Hand dryers are effective in killing the new coronavirus" in description logic\label{fig:m6}}
\end{figure}

Let the myth "Hand dryers are effective in killing the new coronavirus", whose automatic translation in DL appears in Figure~\ref{fig:m6}. 
Fred creates the individual $situation_1:Situation$. 
The role $involves$ from the boxing ontology is used to relate $situation_1$ with the instance $hand\_dryers$:
\begin{align}
boxing:involves(\co situation_1,\co hand\_dryers_1) 
\end{align}
Note that $hand\_dryers_1$ is an instance of the concept $Hand\_dryer$ from the DBpedia. 
The plural is formalised by the role $hasQuantifier$ from the $Quant$ ontology:
\begin{align}
\hspace{-0cm}\quant hasQuantifier(\co hand\_dryers_1, \quant multiple)
\end{align}
The information that hand dryers are effective is modeled with the role $hasQuality$ from the $Dolce$ ontology:
\begin{equation}
    \hspace{-0cm}dul:hasQuality(\co hand\_dryers_1,\co effective)
\end{equation}
Note also that the instance $effective$ is related to the instance $situation_1$ with the role $involves$:
\begin{equation}
 boxing:involves(\co situation_1,\co effective)
\end{equation}
The instance $kill_1$ is identified as an instance of the $Kill_{42030000}$ verb from the VerbNet and also as an instance of the $Event$ concept from the $Dolce$ ontology:
\begin{eqnarray}
\co kill_1:\co Kill\\
\co Kill \equiv vn.data:Kill_{42030000}\\
\co Kill \sqsubseteq dul:Event
\end{eqnarray}

Fred creates the new complex concept $NewCoronavirus$ that is a subclass of the $Coronavirus$ concept from DBpedia and has quality $New$:
\begin{eqnarray}
\co NewCoronavirus \sqsubseteq dbpedia:Coronavirus\ \sqcap\ dul:hasQuality.(\co New) \\
\co New \sqsubseteq dul:Quality
\end{eqnarray}
Here the concept $New$ is identified as a subclass of the $Quality$ concept from Dolce.

Note that Fred has successfully linked the information from the myth with relevant concepts from DBpedia, Verbnet, or Dolce ontologies. 
It also nicely formalises the plural of "dryers",uses compositional semantics for "hand dryers" and "new coronavirus", 

Here, the instance $kill_1$ has the object $coronavirus_1$ as patient.
(Note that the $Patient$ role has the semantics from the VerbNet ontology 
and there is no connection with the patient as a person suffering from the disease). 
Also the instance $kill_1$ has Agent something (i.e. $thing_1$) to which the $situation_1$ is in:
\begin{eqnarray}
\co in(\co situation_1,\co thing_1)\\
vn.role:Agent(\co kill_1,\co thing_1)\\
vn.role:Patient(\co kill_1,\co coronavirus_1)
\end{eqnarray}
The translating meaning would be: 
"The situation involving hand dryers is in something that kills the new coronavirus".

One possible flaw in the automatic translation from Figure~\ref{fig:m6} is that hand dryers are identified as the same individual as coronavirus:
\begin{equation}
\hspace{-0cm}owl:sameAs(\co hand\_dryers_1,\co coronavirus_1)    
\end{equation}
This might be because the term "are" from the myth ("Hand dryers are ....") which signals a possible definition or equivalence. 
These flaw requires post-processing. For instance, we can automatically remove all the relations $sameAs$ from the generated Abox. 

\begin{figure*}
\begin{center}
\includegraphics[width=\textwidth]{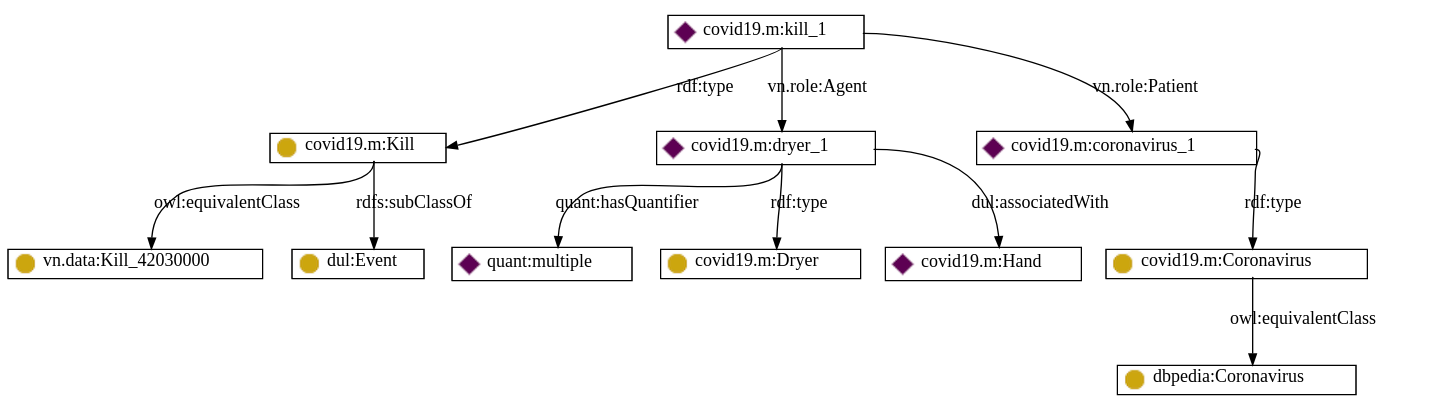}    
\end{center}
\caption{Translating the simplified sentence: "Hand dryers kill coronavirus"\label{fig:m6b}}
\end{figure*}

Actually, the information encapsulated in the given sentence is: 
"Hand dryers kill coronavirus". 
Given this simplified version of the myth, Fred outputs the translation in Figure~\ref{fig:m6b}. 
Here the individual $kill_1$ is correctly linked with the corresponding verb from VerbNet and also identified as an event in Dolce. 
The instance $kill_1$ has the agent $dryer_1$ and the patient $coronavirus_1$. 
This corresponds to the intended semantics: hand dryers kill coronavirus.


\subsection{Modalities and disambiguation}
Deceptive information makes extensively use of modalities. 

Since OWL lacks formal constructs to express modality, FRED uses the Modality class from the Boxing ontology:
\begin{itemize}
    \item \textit{boxing:Necessary}: e.g., will, must, should 
    \item \textit{boxing:Possible}: e.g. may, might
\end{itemize}
where $Necessary \sqsubseteq Modality$ and $Possible \sqsubseteq Modality$

\begin{figure*}
\begin{center}
\includegraphics[width=\textwidth]{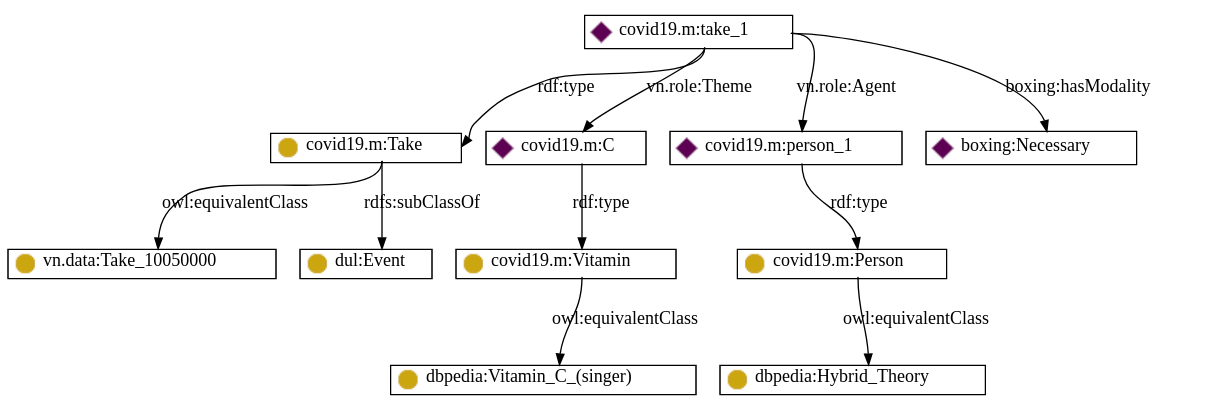}    
\end{center}
\caption{Translating myths with modalities: "You should take vitamin C"\label{fig:m7}}
\end{figure*}

Let the following myth related to Covid-19 "You should take vitamin C" (Figure~\ref{fig:m7}).
The frame is formalised around the instance $take_1$. 
The instance is related to the corresponding verb from the VerbNet and also as an event from the Dolce ontology. 
The agent of the take verb is a person and has the modality $necessary$. 
The individual $C$ is an instance of concept $Vitamin$. 

\begin{figure*}
\begin{center}
\includegraphics[width=\textwidth]{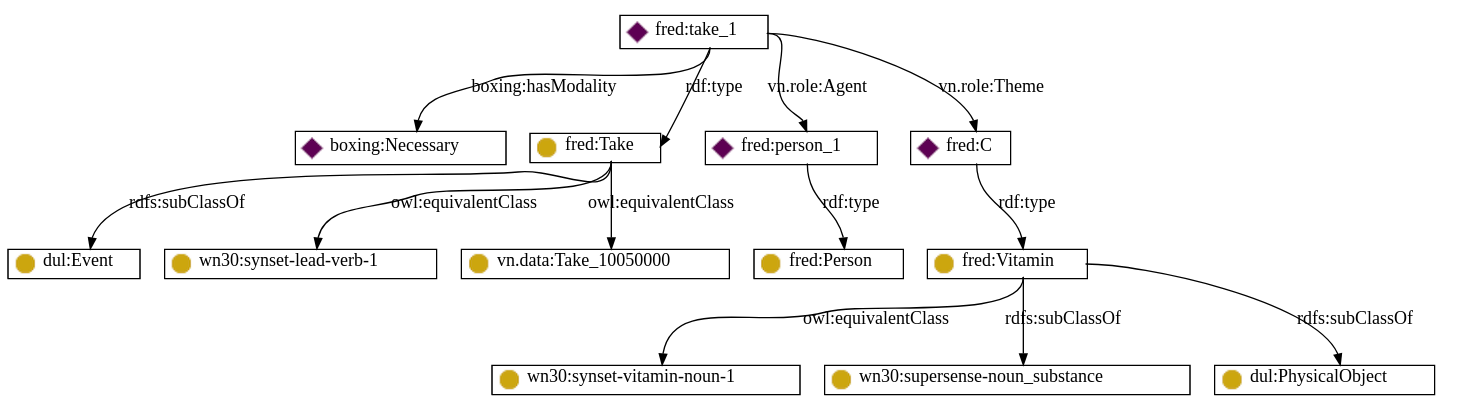}    
\end{center}
\caption{Word sense disambiguation for: "You should take vitamin C"\label{fig:m7b}}
\end{figure*}

Although the above formalisation is correct, the following axioms are wrong.
First, Fred links the concept Vitamin from the Covid-19 ontology with the Vitamin C singer from DBpedia. 
Second, the concept Person from the Covid-19 ontology is linked with Hybrid theory album from the DBpedia, instead of the Person from schema.org.
By performing word sense disambiguation (see Figure~\ref{fig:m7b}, Fred correctly links the vitamin C concept with the noun $vitamin$ from WordNet that is a subclass of the $substance$ concept in the word net and aslo of $PhysicalObject$ from Dolce.

\subsection{Handling negation}

\begin{figure}
\begin{center}
\includegraphics[width=\textwidth]{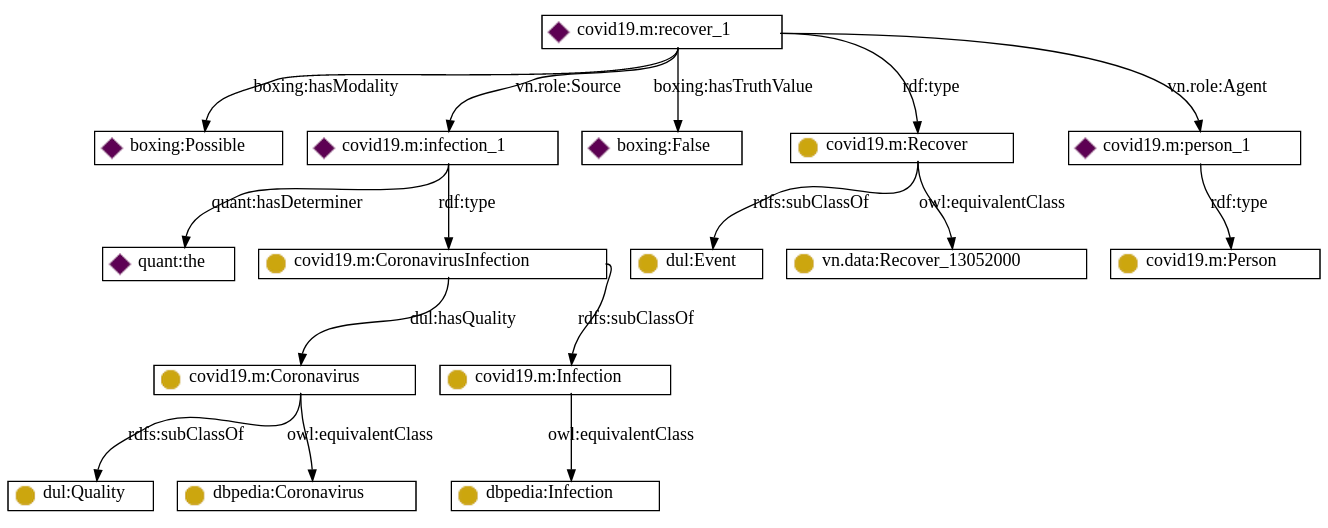}    
\end{center}
\caption{Formalising negations: "You can not recover from the coronavirus infection"\label{fig:m8b}}
\end{figure}

Most of the myths are in positive form. 
For instance, in Table~\ref{tab:myths} only myths $m_3$ and $m_{22}$ includes negation.
Let the translation of myth $m_3$ in Figure~\ref{fig:m8b}.
The frame is built around the $recover_1$ event ($recover_1$ is an instance of $dul:event$ concept) 
Indeed, \fred signals that the event $recover_1$:
\begin{itemize}
    \item has truth value false (axiom~\ref{eq:rec1})
    \item has modality "possible" (axiom~\ref{eq:rec2})
        \item has agent a person (axiom~\ref{eq:rec3})
    \item has source an infection of type coronavirus (axiom~\ref{eq:rec4})
\end{itemize}

\begin{eqnarray}
\hspace{-0cm}boxing:hasTruthValue(\co recover_1,boxing:False) \label{eq:rec1}\\
\hspace{-0cm}boxing:hasModality(\co recover_1,boxing:Possible)\label{eq:rec2}\\
\hspace{-0cm}vn.role:Agent(\co recover_1,\co person_1)\label{eq:rec3}\\
\hspace{-0cm}vn.role:Source(\co recover_1,\co infection_1)\label{eq:rec4}\\
\hspace{-0cm}\co infection_1:\co CoronavirusInfection\\
\hspace{-0cm}\co CoronavirusInfection) \sqsubseteq dbpedia:Infection\ \sqcap\ \exists hasQuality.(dbpedia:Coronavirus)
\end{eqnarray}

However, \fred does not make any assumption on the impact of negation over logical quantification and scope. 
The $boxing:false$ is the only element that one can use to signal conflict between positive and negated information.


\section{Fake news detection by reasoning in Description Logics}
\label{sec:patterns}

Given a possible myth $m_i$ automatically translated by Fred into the ontology $\mathcal{M}^{Fred}_i$, we tackle to fake detection task with three  approaches:  
\begin{enumerate}
\item signal conflict between $\mathcal{M}_i$ and scientific facts $f_j$ also automatically translated by Fred $\mathcal{M}^{Fred}_i$
\item signal conflict between $\mathcal{M}_i$ and the \cov ontology designed by the human agent
\end{enumerate}

\subsection{Detecting conflicts between automatic translation of myths and facts}

\begin{enumerate}
    \item Translating the myth $m_i$ in DL using Fred: $\mathcal{M}^{Fred}_i$
    \item  Translating the fact $f_i$ in DL using Fred  $\mathcal{F}^{Fred}_j$
    \item  Merging the two ontologies $\mathcal{M}^{Fred}_i$ and $\mathcal{F}^{Fred}_j$
    \item Checking the coherence and consistency of the merged ontology $\mathcal{MF}^{Fred}_{ij}$ 
    \item If conflict is detected, Verbalise explanations for the inconsistency
    \item If conflict is not detected import relevant knowledge that may signal the conflict
\end{enumerate}

Consider the pair: 
\begin{center}
\begin{tabular}{ll}
Myth $m_{33}$: & \cov can affect elderly only.\\
Fact $f_{33}$: & \cov can affect anyone.
\end{tabular}
\end{center}
\begin{figure*}
\includegraphics[width=\textwidth]{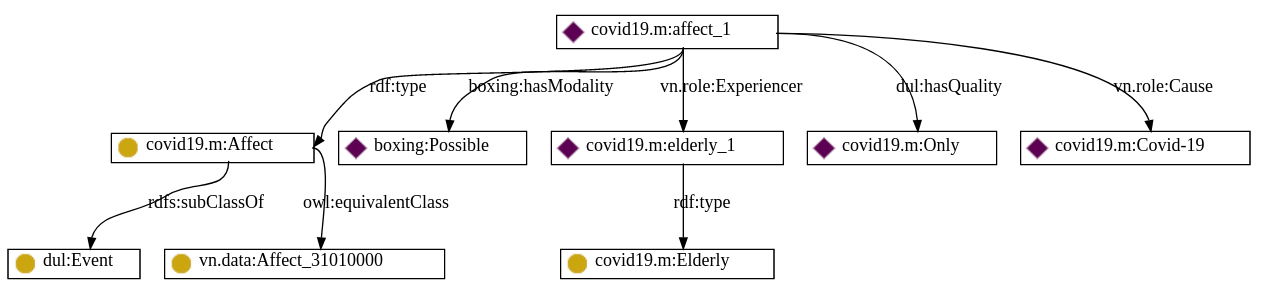}
\caption{Step 1: $\mathcal{M}^{Fred}_{33}$ Automatically translating the myth into DL: ``\cov can affect elderly only``\label{fig:m01}}
\end{figure*}

\begin{figure*}
\includegraphics[width=\textwidth]{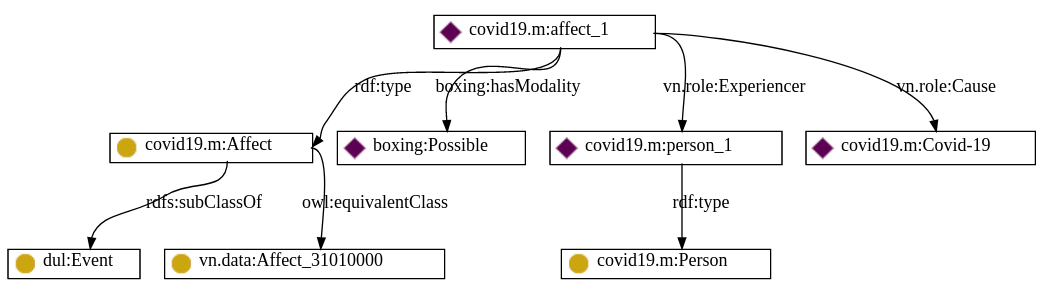}
\caption{Step 2: $\mathcal{F}^{Fred}_{33}$ Automatically translating the fact into DL: ``\cov can affect anyone.``\label{fig:f01}}
\end{figure*}

\begin{figure}
\begin{eqnarray}
Affect \sqsubseteq dul:Event\\
affect_1: Affect\\
elderly_1:Elderly\\
person_1:Person\\
boxing:hasModality(affect_1,boxing:Possible) \\
dul:hasQuality(affect_1, only)\\
vn.role:Cause(affect_1, covid-19)\\
vn.role:Experiencer(affect_1, elderly_1) \\
vn.role:Experiencer(affect_1, person_1)
\end{eqnarray}
\caption{Step 3: Sample of knowledge from the merged ontology $\mathcal{MF}^{Fred}_{(33)(33)}$ \label{fig:step3}} 
\end{figure}
 Figure~\ref{fig:step3} shows the relevant knowledge used to detect conflict (Note that the prefix for the \cov-Myths ontology has been removed).
 The \fred tool has detected the modality $possible$ for the individual $affect_1$. 
 The same instance $affect_1$ has quality $Only$. 
 However, the role $experiencer$ relates the instance $affect_1$ with two individuals: $elderly_1$ and $person_1$. 

\begin{figure}
\begin{eqnarray}
Elderly \sqsubseteq Person\\
person_1: \neg Elderly\\
(?x\ Only\ hasQuality) \wedge (?x\ ?y\ Experiencer) \wedge (?x\ ?z\ Experiencer) \wedge (?y\ Elderly) \nonumber\\
\rightarrow (?z\ Elderly)
\end{eqnarray}
\caption{Step 4: Conflict detection based on the pattern $\exists dul:hasQuality.Only$ \label{fig:step4}} 
\end{figure}
The axioms in Figure~\ref{fig:step4} state that an elderly is a person and that the instance $person_1$ is not elderly.
The conflict detection pattern is defined as: $\exists dul:hasQuality.Only$ 
The SWRL rule states that for each individual $?x$ with the quality $only$ that is related via the role $experiencer$ 
with two distinct individuals $?y$ and $?z$ (where $?y$ is an instance of the concept $Elderly$), then the individual $?z$ is also an instance of $Elderly$.

The conflict comes from the fact that $person_1$ is not an instance of $Elderly$, but still he/she is affected by COVID (i.e. $experiencer(affect_1, person_1)$).






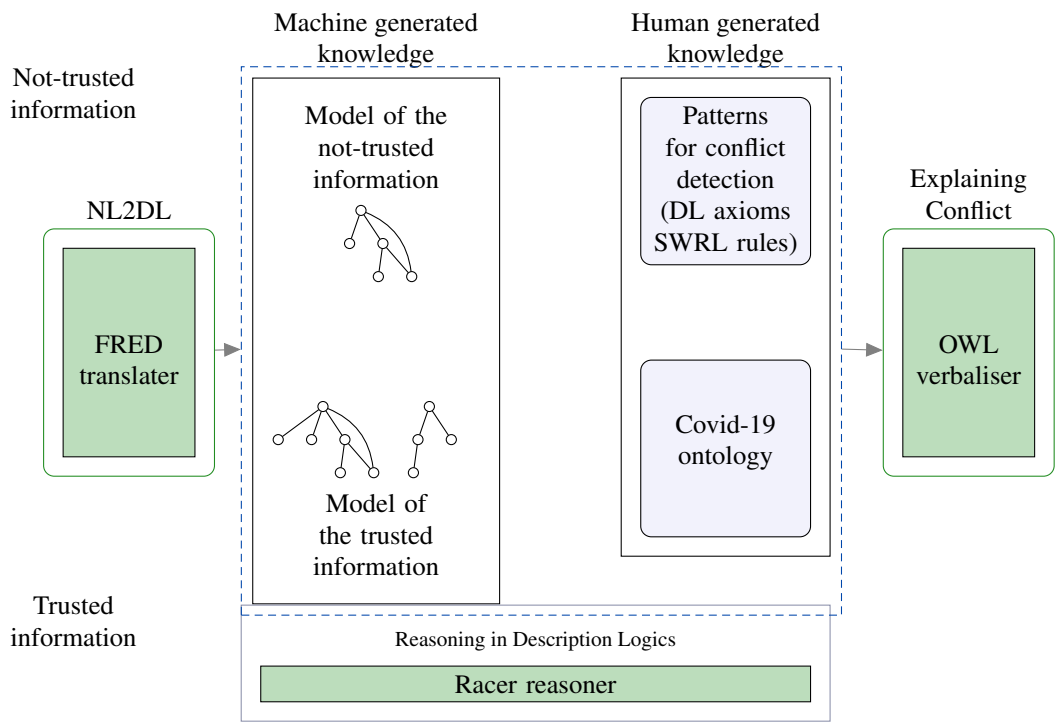
\begin{figure*}
\definecolor{lavenderweb}{rgb}{0.9, 0.9, 0.98}
\definecolor{lavenderblue}{rgb}{0.8, 0.8, 1.0}
\definecolor{forestgreen}{rgb}{0.13, 0.55, 0.13}
\definecolor{emerald}{rgb}{0.31, 0.78, 0.47}
\definecolor{darkpastelblue}{rgb}{0.47, 0.62, 0.8}
\definecolor{coolgrey}{rgb}{0.55, 0.57, 0.67}
\definecolor{cobalt}{rgb}{0.0, 0.28, 0.67}
\definecolor{champagne}{rgb}{0.97, 0.91, 0.81}
\definecolor{bostonuniversityred}{rgb}{0.8, 0.0, 0.0}
\definecolor{languidlavender}{rgb}{0.84, 0.79, 0.87}
\definecolor{orange-red}{rgb}{1.0, 0.27, 0.0}
\definecolor{gray}{rgb}{0.5, 0.5, 0.5}


\tikzstyle{spec}=[rectangle, draw=black, rounded corners, fill=lavenderweb!50, 
        text centered, text width=2cm]
\tikzstyle{ui}=[rectangle, draw=champagne, rounded corners, fill=coolgrey, 
        text centered, anchor=north, text=champagne, text width=5cm, node distance=0.3cm]
\tikzstyle{kbnew}=[circle, draw=black, fill=white, text centered, anchor=north, text=black, text width=0.005cm, node distance=0.3cm]
\tikzstyle{trust}=[rectangle, draw=black,  fill=forestgreen, 
        text centered, anchor=north, text=black, text width=1.3cm, node distance=0.2cm]
\tikzstyle{limitedtrust}=[rectangle, draw=black,  fill=forestgreen!30,text centered, anchor=north, text=black, text width=1.5cm, node distance=0.2cm]
\tikzstyle{limitedtrustlarge}=[rectangle, draw=black,  fill=forestgreen!30,text centered, anchor=north, text=black, text width=7cm, node distance=0.2cm]
\tikzstyle{nottrust}=[rectangle, draw=champagne, rounded corners, fill=champagne!60, 
        text centered, anchor=north, text=bostonuniversityred, text width=3cm, node distance=0.2cm]
\tikzstyle{onlytext}=[text centered, anchor=north, text=black, text width=2.5cm, node distance=0.1cm]
\tikzstyle{uionlytext}=[text centered, anchor=north, text=black,  node distance=0.1cm]        
\tikzstyle{myarrow}=[->, >=triangle 45, color=gray ]

\tikzstyle{line}=[-, thick, draw=red, inner sep=2mm]


\begin{tikzpicture}
    
    \node(nottrusted)[onlytext, rectangle, inner sep=2.5mm, xshift=-8cm, label={[align=center]above:Not-trusted \\ information}] {};
    \node(trusted)[onlytext, rectangle, inner sep=2.5mm, xshift=-8cm, yshift=-7cm,label={[align=center]above:Trusted \\ information}] {};
    
    \node (modelnews) [onlytext, rectangle, right=of nottrusted, xshift=1cm] {Model  of the not-trusted information};
    \node (first) [kbnew, below=of modelnews, scale=0.4, yshift=0.5cm, xshift=-0.5cm] {};
    \node (second) [kbnew, below=of first, xshift=-0.15cm, scale=0.4] {};
    \node (third) [kbnew, right=of second , scale=0.4] {};
    \node (fourth) [kbnew, below=of third , scale=0.4, xshift=-0.15cm,] {};
    \node (fifth) [kbnew, right=of fourth , scale=0.4,] {};
   
    
    \node (first1) [kbnew, below=of modelnews, yshift=-2.4cm,xshift=-0.7cm, scale=0.4] {};
    \node (second1) [kbnew, below=of first1, xshift=-0.15cm, scale=0.4] {};
    \node (third1) [kbnew, right=of second1 , scale=0.4] {};
    \node (fourth1) [kbnew, below=of third1 , scale=0.4, xshift=-0.15cm,] {};
    \node (fifth1) [kbnew, right=of fourth1 , scale=0.4,] {};
    \node (sixth1) [kbnew, left=of second1 , scale=0.4,] {};
    
    \node (first2) [kbnew, below=of modelnews, yshift=-2.4cm,xshift=0.7cm, scale=0.4] {};
    \node (second2) [kbnew, below=of first2, xshift=-0.15cm, scale=0.4] {};
    \node (third2) [kbnew, right=of second2 , scale=0.4] {};
    \node (fourth2) [kbnew, below=of second2 , scale=0.4, xshift=-0.15cm,] {};
    
     \node (modeltrusted) [onlytext, rectangle, below=of modelnews, yshift=-3.7cm] {Model of the trusted information};
    
    \node(secondrow)[draw,inner sep=2.5mm,label={[align=center]above:Machine generated \\ knowledge},fit=(modelnews) (modeltrusted) (modelnews) (modelnews)] {};

     \node (specification) [spec, right=of modelnews.north east, anchor=north west, xshift=1.1cm] {Patterns\\for conflict \\detection \\(DL axioms \\SWRL rules) \\   };
      \node (specification1) [spec, below=of specification, yshift=-0.25cm] {\quad \\ \quad \\ Covid-19 ontology \\\quad \\\quad \\   };
      
     
     \node(thirdrow)[draw=black,inner sep=2.5mm,label={[align=center]above:Human generated \\knowledge},fit=(specification) (specification1) (specification) (specification)] {};
     
     \node(onto)[limitedtrust, right=of specification.north east,anchor=north west, xshift=1cm, yshift=-2cm] {\\\quad \\\quad \\OWL\\ verbaliser \\\quad \\\quad \\};
     \node(fourthrow)[draw=forestgreen,rounded corners, inner sep=2.5mm,label={[align=center]above:Explaining \\ Conflict},fit=(onto)] {};

    \node (fred) [limitedtrust, left=of modelnews.north east,anchor=north west, xshift=-5.3cm, yshift=-2cm] {\\\quad \\\quad \\FRED\\ translater \\\quad \\\quad \\};
    \node(fred1)[draw=forestgreen,rounded corners, inner sep=2.5mm,label={[align=center]above:NL2DL},fit=(fred)] {};
    
     \node (arg) [limitedtrustlarge, below=of specification1,xshift=-2.5cm, yshift=-1.5cm] {Racer reasoner};
     \node (chat) [uionlytext, above=of arg,scale=0.8] {Reasoning in Description Logics};
     \node(down)[draw=coolgrey, inner sep=2.5mm,label={[align=center, text=coolgrey]left:},fit=(chat) (arg)] {};
    
    \node(core)[draw=cobalt, densely dashed,  inner sep=4mm,fit=(modelnews) (modeltrusted) (modelnews) (specification)] {};
    
    
   
 
   \draw[myarrow] (core) edge node[above, scale=0.8]{} (fourthrow);
   \draw[myarrow] (fred1) edge node[right, scale=0.8]{} (core);
   
   \draw[thin] (first) -- (second);
   \draw[thin] (first) -- (third);
   \draw[thin] (third) -- (fourth);
   \draw[thin] (third) -- (fifth);
   \draw[thin] (first) to [bend left=30] (fifth);
   
   \draw[thin] (first1) -- (second1);
   \draw[thin] (first1) -- (third1);
   \draw[thin] (third1) -- (fourth1);
   \draw[thin] (third1) -- (fifth1);
   \draw[thin] (first1) -- (sixth1);
   \draw[thin] (first1) to [bend left=30] (fifth1);
   
   \draw[thin] (first2) -- (second2);
   \draw[thin] (first2) -- (third2);
   \draw[thin] (second2) -- (fourth2);
    
\end{tikzpicture} 
\caption{A \cov ontology is enriched using FRED with trusted facts and medical myths. Racer reasoner is used to detect inconsitencies in the enriched ontology, based on some patterns manually formalised in Description Logics or SWRL
\label{fig:arch}}
\end{figure*}

The system architecture appears in Figure~\ref{fig:arch}. 
We start with a core ontology for \cov. 
This ontology is enriched with trusted facts on COVID using the FRED converter. 
Information  from untrusted sources is also formalised in DL using FRED. 
The merged axioms are given to Racer that is able to signal conflicts.

To support the user understanding which knowledge from the ontology is causing incoherences, we use the Racer's explanation capabilities. 
RacerPro provides explanations for unsatisfiable concepts, for subsumption relationships, and for unsatisfiable A-boxes through the commands \textit{(check-abox-coherence)}, \textit{(check-tbox-coherence)} and \textit{(check-ontology)} or \textit{(retrieve-with-explanation)}. 
These explanations are given to an ontology verbalizer in order to generated natural language explanation of the conflict.


We aim to collect a corpus of common misconceptions that are spread in online media. 
We aim to analyse these misconceptions and to build evidence-based counter arguments to each piece of deceptive information.
We aim to annotate this corpus with concepts and roles from trusted medical ontologies.

\section{Discussion and related work}
\label{sec:related}
Our topic is related to the more general issue of fake news~\cite{FIGUEIRA2017817}.
Particular to medical domain, there has been a continuous concern of reliability of online heath information~\cite{ADAMS2010391}.
In this line, Waszak et al. have recently investigated the spread of fake medical news in social media~\cite{waszak2018spread}.
Amith and Tao have formalised the Vaccine Misinformation Ontology (VAXMO)~\cite{amith2018representing}. 
VAXMO extends the Misinformation Ontology, aiming to support vaccine misinformation detection and analysis\footnote{http://www.violinet.org/vaccineontology/}.

Teymourlouie et al. have recently analyse the importance of contextual knowledge in detecting ontology conflicts. 
The added contextual knowledge is applied in~\cite{teymourlouie2018detecting}\footnote{https://github.com/teymourlouie/ontodebugger} to the task fo debugging ontologies. 
In our case, the contextual ontology is represented by patterns of conflict detection between two merged ontologies. 
The output of \fred is given to the Racer reasoner that detects conflict based 
on trusted medical source and conflict detection patterns.


FiB system~\cite{figueira2017current} labels news as verified or non-verified. 
It crawls the Web for similar news to the current one and summarised them  
The user reads the summaries and figures out which information from the initial new might be fake.
 We aim a step forward, towards automatically identify possible inconsistencies between a given news and the verified medical content. 




MERGILO tool reconciles knowledge graphs extracted from text, using graph alignment and word similarity~\cite{alam2017event}. 
One application area is to detect knowledge evolution across document versions. To obtain the formalisation of events, MERGILO used both \fred and Framester. Instead of using metrics for compute graph similarity, I used here knowledge patterns to detect conflict.  

Enriching ontologies with complex axioms has been given some consideration in literature~\cite{georgiu2011ontology,gyawali2017mapping}. 
The aim would be bridge the gap between a document-centric and a model-centric view of information~\cite{gyawali2017mapping}.
Gyawali et al translate text in the SIDP format (i.e. System Installation Design Principle) to axioms in description logic.
The proposed system combines an automatically derived lexicon with a hand-written grammar to automatically generates axioms. 
Here, the core \cov ontology is enriched with axioms generated by Fred fed with facts in natural language. 
Instead of grammar, I formalised knowledge patterns (e.g. axioms in DL or SWRL rules) to detect conflicts. 

Conflict detection depends heavily on the performance of the FRED translator One can replace FRED by related tools such as Framester~\cite{gangemi2016framester} or KNEWS~\cite{basile2016knews}. 
Framester is a large RDF knowledge graph (about 30 million RDF triples) acting as a umbrella for FrameNet, WordNet, VerbNet, BabelNet, Predicate Matrix. 
In contrast to FRED, KNEWS (Knowledge Extraction With Semantics) can be configured to use different external modules a
s input, but also different output modes (i.e. frame instances, word aligned semantics or first order logic\footnote{https://github.com/valeriobasile/learningbyreading}). 
Frame representation outputs RDF tuples in line with the FrameBase\footnote{http://www.framebase.org/} model. 
First-order logic formulae in syntax similar to TPTP and they include WordNet synsets and DBpedia ids as symbols~\cite{basile2016knews}.

\section{Conclusion}
\label{sec:conclusion}
Even if fake news in the health domain is old hat, many technical challenges 
remain to effective fight against medical myths.  
This is preliminary work on combining two heavy machineries: natural language processing and ontology reasoning aiming to signal fake information related to \cov.  

The ongoing work includes: i) system evaluation and ii) verbalising explanations for each identified conflict.
\bibliographystyle{plain}
\bibliography{main} 

\begin{thebibliography}{10}

\bibitem{ADAMS2010391}
Samantha~A. Adams.
\newblock Revisiting the online health information reliability debate in the
  wake of web 2.0: An inter-disciplinary literature and website review.
\newblock {\em International Journal of Medical Informatics}, 79(6):391 -- 400,
  2010.
\newblock Special Issue: Information Technology in Health Care: Socio-technical
  Approaches.

\bibitem{alam2017event}
Mehwish Alam, Diego~Reforgiato Recupero, Misael Mongiovi, Aldo Gangemi, and
  Petar Ristoski.
\newblock Event-based knowledge reconciliation using frame embeddings and frame
  similarity.
\newblock {\em Knowledge-Based Systems}, 135:192--203, 2017.

\bibitem{amith2018representing}
Muhammad Amith and Cui Tao.
\newblock Representing vaccine misinformation using ontologies.
\newblock {\em Journal of biomedical semantics}, 9(1):22, 2018.

\bibitem{baader2003description}
Franz Baader, Diego Calvanese, Deborah McGuinness, Peter Patel-Schneider,
  Daniele Nardi, et~al.
\newblock {\em The description logic handbook: Theory, implementation and
  applications}.
\newblock Cambridge university press, 2003.

\bibitem{baker1998berkeley}
Collin~F Baker, Charles~J Fillmore, and John~B Lowe.
\newblock The berkeley framenet project.
\newblock In {\em Proceedings of the 17th international conference on
  Computational linguistics-Volume 1}, pages 86--90. Association for
  Computational Linguistics, 1998.

\bibitem{basile2016knews}
Valerio Basile, Elena Cabrio, and Claudia Schon.
\newblock Knews: Using logical and lexical semantics to extract knowledge from
  natural language.
\newblock 2016.

\bibitem{cinelli2020covid}
Matteo Cinelli, Walter Quattrociocchi, Alessandro Galeazzi, Carlo~Michele
  Valensise, Emanuele Brugnoli, Ana~Lucia Schmidt, Paola Zola, Fabiana Zollo,
  and Antonio Scala.
\newblock The covid-19 social media infodemic.
\newblock {\em arXiv preprint arXiv:2003.05004}, 2020.

\bibitem{draicchio2013fred}
Francesco Draicchio, Aldo Gangemi, Valentina Presutti, and Andrea~Giovanni
  Nuzzolese.
\newblock Fred: From natural language text to rdf and owl in one click.
\newblock In {\em Extended Semantic Web Conference}, pages 263--267. Springer,
  2013.

\bibitem{FIGUEIRA2017817}
Alvaro Figueira and Luciana Oliveira.
\newblock The current state of fake news: challenges and opportunities.
\newblock {\em Procedia Computer Science}, 121:817 -- 825, 2017.

\bibitem{figueira2017current}
Alvaro Figueira and Luciana Oliveira.
\newblock The current state of fake news: challenges and opportunities.
\newblock {\em Procedia Computer Science}, 121:817--825, 2017.

\bibitem{gangemi2016framester}
Aldo Gangemi, Mehwish Alam, Luigi Asprino, Valentina Presutti, and
  Diego~Reforgiato Recupero.
\newblock Framester: a wide coverage linguistic linked data hub.
\newblock In {\em European Knowledge Acquisition Workshop}, pages 239--254.
  Springer, 2016.

\bibitem{gangemi2017semantic}
Aldo Gangemi, Valentina Presutti, Diego Reforgiato~Recupero, Andrea~Giovanni
  Nuzzolese, Francesco Draicchio, and Misael Mongiov{\`\i}.
\newblock Semantic web machine reading with fred.
\newblock {\em Semantic Web}, 8(6):873--893, 2017.

\bibitem{georgiu2011ontology}
Marius Georgiu and Adrian Groza.
\newblock Ontology enrichment using semantic wikis and design patterns.
\newblock {\em Studia Universitatis Babes-Bolyai, Informatica}, 56(2):31, 2011.

\bibitem{gyawali2017mapping}
Bikash Gyawali, Anastasia Shimorina, Claire Gardent, Samuel Cruz-Lara, and
  Mariem Mahfoudh.
\newblock Mapping natural language to description logic.
\newblock In {\em European Semantic Web Conference}, pages 273--288. Springer,
  2017.

\bibitem{haarslev2012racerpro}
Volker Haarslev, Kay Hidde, Ralf M{\"o}ller, and Michael Wessel.
\newblock The racerpro knowledge representation and reasoning system.
\newblock {\em Semantic Web}, 3(3):267--277, 2012.

\bibitem{lazer2018science}
David~MJ Lazer, Matthew~A Baum, Yochai Benkler, Adam~J Berinsky, Kelly~M
  Greenhill, Filippo Menczer, Miriam~J Metzger, Brendan Nyhan, Gordon
  Pennycook, David Rothschild, et~al.
\newblock The science of fake news.
\newblock {\em Science}, 359(6380):1094--1096, 2018.

\bibitem{martinez2018openie}
Jose~L Martinez-Rodriguez, Ivan Lopez-Arevalo, and Ana~B Rios-Alvarado.
\newblock Openie-based approach for knowledge graph construction from text.
\newblock {\em Expert Systems with Applications}, 113:339--355, 2018.

\bibitem{roussey2013antipattern}
Catherine Roussey and A~Zamazal.
\newblock Antipattern detection: how to debug an ontology without a reasoner.
\newblock 2013.

\bibitem{schuler2005verbnet}
Karin~Kipper Schuler.
\newblock Verbnet: A broad-coverage, comprehensive verb lexicon.
\newblock 2005.

\bibitem{teymourlouie2018detecting}
Mehdi Teymourlouie, Ahmad Zaeri, Mohammadali Nematbakhsh, Matthias Thimm, and
  Steffen Staab.
\newblock Detecting hidden errors in an ontology using contextual knowledge.
\newblock {\em Expert Systems with Applications}, 95:312--323, 2018.

\bibitem{vosoughi2018spread}
Soroush Vosoughi, Deb Roy, and Sinan Aral.
\newblock The spread of true and false news online.
\newblock {\em Science}, 359(6380):1146--1151, 2018.

\bibitem{waszak2018spread}
Przemyslaw~M Waszak, Wioleta Kasprzycka-Waszak, and Alicja Kubanek.
\newblock The spread of medical fake news in social media--the pilot
  quantitative study.
\newblock {\em Health Policy and Technology}, 2018.

\end{thebibliography}

\end{document}